\pdfoutput=1
\documentclass[11pt]{article}
\usepackage{multirow}

\usepackage[table,dvipsnames]{xcolor}

\usepackage[]{naacl2021}

\usepackage{times}
\usepackage{latexsym}

\usepackage[T1]{fontenc}
\usepackage[utf8]{inputenc}

\usepackage{microtype}
\usepackage{graphicx}
\usepackage{amsmath}

\usepackage[ruled,vlined]{algorithm2e}
\DeclareMathAlphabet{\pazocal}{OMS}{zplm}{m}{n}

\usepackage{microtype}

\usepackage{graphicx}               % Include graph
\usepackage{tabularx}               % Better table formatting
\newcolumntype{C}{>{\centering\arraybackslash}X}
\usepackage{multirow}               % Multi-row in tables
\usepackage{diagbox}                % Diagonal line in tables
\usepackage{hhline}                 % Draw double line in tables
\usepackage{color}                  % Text and background color
\usepackage{amssymb}                % Formula
\usepackage{mathtools}              % Formula
\usepackage{enumitem}               % Better itemize environment

\usepackage{makecell}

\DeclareMathOperator*{\argmax}{arg\,max}

\usepackage{xparse}
\NewDocumentCommand{\heng}{ mO{} }{\textcolor{OrangeRed}{\textsuperscript{\textit{Heng}}\textsf{\textbf{\small[#1]}}}}
\NewDocumentCommand{\tuan}{ mO{} }{\textcolor{Blue}{\textsuperscript{\textit{Tuan}}\textsf{\textbf{\small[#1]}}}}

\title{A Context-Dependent Gated Module for Incorporating Symbolic Semantics into Event Coreference Resolution}

\author{Tuan Lai, Heng Ji \\
	    University of Illinois at Urbana-Champaign\\
\{tuanml2, hengji\}@illinois.edu\\ 
\textbf{Trung Bui, Quan Tran, Franck Dernoncourt, Walter Chang}\\
	    Adobe Research\\
        \{bui, qtran, franck.dernoncourt, wachang\}@adobe.com
}

\begin{document}
\maketitle

\begin{abstract}

Event coreference resolution is an important research problem with many applications. Despite the recent remarkable success of pre-trained language models, we argue that it is still highly beneficial to utilize symbolic features for the task. However, as the input for coreference resolution typically comes from upstream components in the information extraction pipeline, the automatically extracted symbolic features can be noisy and contain errors. Also, depending on the specific context, some features can be more informative than others. Motivated by these observations, we propose a novel context-dependent gated module to adaptively control the information flows from the input symbolic features. Combined with a simple noisy training method, our best models achieve state-of-the-art results on two datasets: ACE 2005 and KBP 2016.\footnote{The code is publicly available at \url{https://github.com/laituan245/eventcoref}.}

% Event coreference is an important research problem with applications in many tasks. Despite the recent remarkable success of pretrained language models, we argue that it is still beneficial to utilize symbolic features when resolving coreferential event mentions. As the inputs for event coreference typically come from upstream components in the information extraction pipeline, the input symbolic features can be noisy and contain errors. In addition, depending on the specific context, some symbolic features can be more informative than others. Motivated by these observations, we propose a novel context-dependent gated module to adaptively control the information flows from the input symbolic features. Furthermore, to encourage our module to actually learn to distill reliable symbolic semantics, we propose a simple regularization method that intentionally adds more noise to the symbolic features during training. Extensive experimental results on both within-document and cross-document coreference benchmarks demonstrate the effectiveness of our proposed methods. Our best models outperform previous methods, achieving new state-of-the-art results on three standard event coreference datasets: ACE-2005, KBP 2016, and ECB+.
\end{abstract}

\section{Introduction}

Within-document event coreference resolution is the task of clustering event mentions in a text that refer to the same real-world events \cite{Lu2018EventCR}. It is an important research problem, with many applications \cite{Vanderwende2004EventCentricSG,jigrishman2011knowledge,Choubey2018IdentifyingTM}. Since the trigger of an event mention is typically the word or phrase that most clearly describes the event, virtually all previous approaches employ features related to event triggers in one form or another. To achieve better performance, many methods also need to use a variety of additional symbolic features such as event types, attributes, and arguments \cite{chenetal2009pairwise,chenji2009graph,zhang2015crossdocument,sammons2015illinois,Lu2016EventCR,Chen2016JointIO,duncan2017ui}. Previous neural methods \cite{nguyen2016new,Choubey2017EventCR,huangetal2019improving} also use non-contextual word embeddings such as word2vec \cite{word2vec} or GloVe \cite{penningtonetal2014glove}.

\renewcommand{\arraystretch}{1.25}
\begin{table}[t!]
\centering
\small
\begin{tabular}{p{0.92\linewidth}}
\hline
... we are seeing these soldiers \textcolor{red}{$\{$head out$\}_{\text{ev1}}$} ... \\
... these soldiers \textbf{were set to} \textcolor{blue}{$\{$leave$\}_{\text{ev2}}$} in January ... \\
\textcolor{red}{\text{ev1}} (\texttt{Movement:Transport}): Modality = ASSERTED\\
\textcolor{blue}{\text{ev2}} (\texttt{Movement:Transport}): Modality = OTHER\\
\hline
\end{tabular}
\caption{\label{table:example-spanbert-error} An example of using the modality attribute to improve event coreference resolution.}
\vspace{-6mm}
\end{table}

With the recent remarkable success of language models such as BERT \cite{devlinetal2019bert} and SpanBERT \cite{Joshi2019SpanBERTIP}, one natural question is whether we can simply use these models for coreference resolution without relying on any additional features. We argue that it is still highly beneficial to utilize symbolic features, especially when they are clean and have complementary information. Table \ref{table:example-spanbert-error} shows an example in the ACE 2005 dataset, where our baseline SpanBERT model incorrectly predicts the highlighted event mentions to be coreferential. The event triggers are semantically similar, making it challenging for our model to distinguish. However, notice that the event \textcolor{red}{$\{$head out$\}_{\text{ev1}}$} is mentioned as if it was a real occurrence, and so its modality attribute is ASSERTED \cite{ace2005english}. In contrast, because of the phrase ``were set to'', we can infer that the event \textcolor{blue}{$\{$leave$\}_{\text{ev2}}$} did not actually happen (i.e., its modality attribute is OTHER). Therefore, our model should be able to avoid the mistake if it utilizes additional symbolic features such as the modality attribute in this case.

There are several previous methods that use contextual embeddings together with type-based or argument-based information \cite{Lu2020EndtoEndNE,Yu2020PairedRL}. For example, \newcite{Lu2020EndtoEndNE} proposes a new mechanism to better exploit event type information for coreference resolution. Despite their impressive performance, these methods are specific to one particular type of additional information.

In this paper, we propose general and effective methods for incorporating a wide range of symbolic features into event coreference resolution. Simply concatenating symbolic features with contextual embeddings is not optimal, since the features can be noisy and contain errors. Also, depending on the context, some features can be more informative than others. Therefore, we design a novel context-dependent gated module to extract information from the symbolic features selectively. Combined with a simple regularization method that randomly adds noise into the features during training, our best models achieve state-of-the-art results on ACE 2005 \cite{walker2006ace} and KBP 2016 \cite{mitamura2016overview} datasets. To the best of our knowledge, our work is the first to explicitly focus on dealing with various noisy symbolic features for event coreference resolution.

%  on ACE 2005 \cite{walker2006ace} and KBP 2016 \cite{mitamura2016overview}
\section{Methods}
\begin{figure*}[!ht]
\centering
\includegraphics[width=0.85\textwidth]{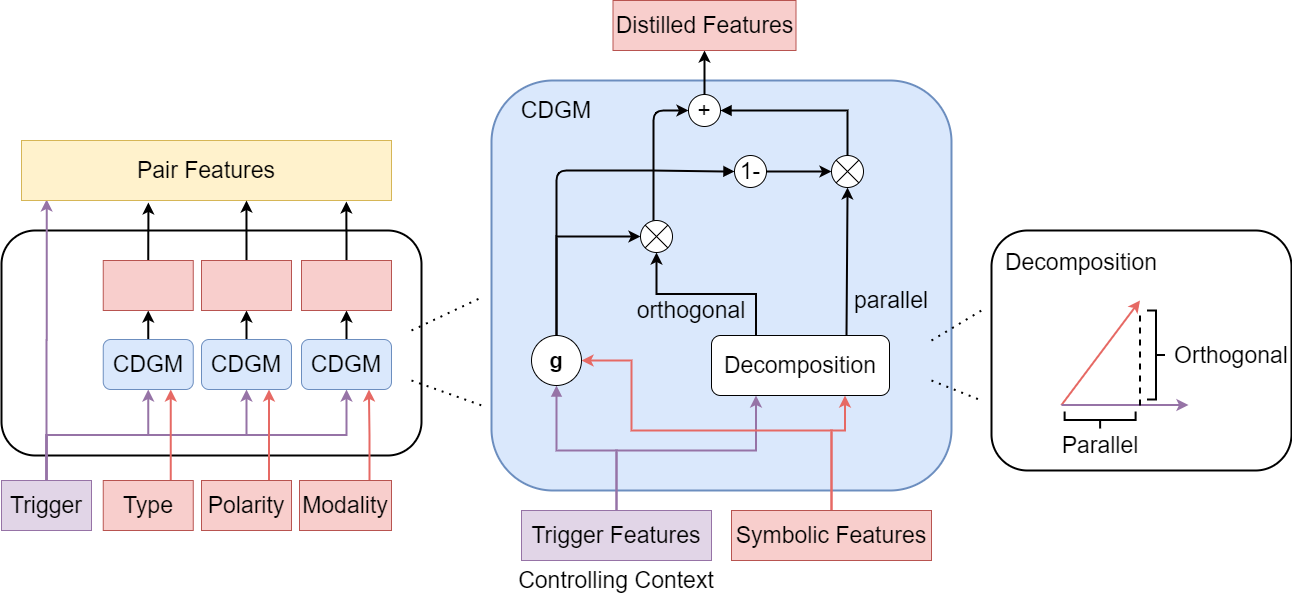}
\caption{Overall architecture of our mention-pair encoder, which uses CDGMs to incorporate symbolic features.}
\label{fig:overall_pair_encoder}
\end{figure*}
\subsection{Preliminaries}\label{sec:prelim}
We focus on within-document event coreference resolution. The input to our model is a document $D$ consisting of $n$ tokens and $k$ (predicted) event mentions $\{m_1, m_2, \dots, m_k\}$. For each $m_i$, we denote the start and end indices of its trigger by $s_i$ and $e_i$ respectively. We assume the mentions are ordered based on $s_i$ (i.e., If $i \leq j$ then $s_i \leq s_j$).

We also assume each $m_i$ has $K$ (predicted) categorical features $\{c_{i}^{(1)}, c_{i}^{(2)}, \dots, c_{i}^{(K)}\}$, with each $c_{i}^{(u)} \in \{1, 2, \dots, N_u\}$ taking one of $N_u$ different discrete values. Table \ref{tab:attributes_list} lists the symbolic features we consider in this work. The definitions of the features and their possible values are in ACE and Rich ERE guidelines \cite{ace2005english,mitamura2016overview}. The accuracy scores of the symbolic feature predictors are also shown in Table \ref{tab:attributes_list}. We use OneIE \cite{linetal2020joint} to identify event mentions along with their subtypes. For other symbolic features, we train a joint classification model based on SpanBERT. The appendix contains more details.

% Without loss of generality
% $(t_1, t_2, \dots, t_n)$
% $\{m_1, m_2, \dots, m_k\}$

% In this work, we focus on within-document event coreference. 
% \cite{Joshi2019SpanBERTIP}
\subsection{Single-Mention Encoder}
Given a document $D$, our model first forms a contextualized representation for each input token using a Transformer encoder \cite{Joshi2019SpanBERTIP}. Let $\textbf{X} = (\textbf{x}_1, ..., \textbf{x}_n)$ be the output of the encoder, where $\textbf{x}_i \in \mathbb{R}^{d}$. Then, for each mention $m_i$, its trigger's representation $\textbf{t}_i$ is defined as the average of its token embeddings:
\begin{equation}
    \textbf{t}_i = \sum \limits_{j=s_i}^{e_i} \frac{\textbf{x}_j}{e_i - s_i + 1}
\end{equation}
Next, by using $K$ trainable embedding matrices, we convert the symbolic features of $m_i$ into $K$ vectors $\{\textbf{h}_{i}^{(1)}, \textbf{h}_{i}^{(2)}, \dots, \textbf{h}_{i}^{(K)}\}$, where $\textbf{h}_{i}^{(u)} \in \mathbb{R}^{l}$. 

% Since the trigger of an event mention typically consists of very few tokens, we simply use the averaging strategy. In ACE 2005, only 4.78\% of the event triggers are multi-word expressions.

%These $K$ feature vectors and the trigger vector $\textbf{t}_i$ collectively contain a lot of information about $m_i$.

%In our preliminary experiments, we also find that using attention does not improve performance but brings extra computational costs.
\begin{table}[!t]
\centering
\resizebox{0.80\linewidth}{!}{%
\begin{tabular}{|c|l|c|c|c|}
\hline
Dataset & Features & \begin{tabular}[c]{@{}c@{}}Acc.\\ (Train)\end{tabular} & \begin{tabular}[c]{@{}c@{}}Acc.\\ (Dev)\end{tabular}  & \begin{tabular}[c]{@{}c@{}}Acc.\\ (Test)\end{tabular} \\ \hline
\multirow{5}{*}{\begin{tabular}[c]{@{}c@{}}ACE\\ 2005\end{tabular}} & Type & 0.999 & 0.945 & 0.953 \\ \cline{2-5} 
 & Polarity & 0.999 & 0.994 & 0.988 \\ \cline{2-5} 
 & Modality & 0.999 & 0.856 & 0.884 \\ \cline{2-5} 
 & Genericity & 0.999 & 0.865 & 0.872 \\ \cline{2-5} 
 & Tense & 0.984 & 0.802 & 0.763 \\ \hline
\multirow{2}{*}{\begin{tabular}[c]{@{}c@{}}KBP\\ 2016\end{tabular}} & Type & 0.960 & 0.874 & 0.818 \\ \cline{2-5} 
 & Realis & 0.979 & 0.845 & 0.840 \\ \hline
\end{tabular}%
}
\caption{Symbolic features we consider in this work.}
\label{tab:attributes_list}
\vspace{-3.5mm}
\end{table}
\subsection{Mention-Pair Encoder and Scorer}
Given two event mentions $m_i$ and $m_j$, we define their trigger-based pair representation as:
\begin{equation}\label{equa:trigger_pair}
\textbf{t}_{ij} = \text{FFNN}_\text{t}\big(\big\lbrack \textbf{t}_i,\textbf{t}_j, \textbf{t}_i \circ \textbf{t}_j \big\rbrack\big)
\end{equation}
where $\text{FFNN}_\text{t}$ is a feedforward network mapping from $\mathbb{R}^{3 \times d} \rightarrow \mathbb{R}^p$, and $\circ$ is element-wise multiplication. Similarly, we can compute their feature-based pair representations $\{\textbf{h}_{ij}^{(1)}, \textbf{h}_{ij}^{(2)}, \dots, \textbf{h}_{ij}^{(K)}\}$ as follows:
\begin{equation}
\textbf{h}_{ij}^{(u)} = \text{FFNN}_\text{u}\big(\big\lbrack \textbf{h}_i^{(u)},\textbf{h}_j^{(u)}, \textbf{h}_i^{(u)} \circ \textbf{h}_j^{(u)} \big\rbrack\big)
\end{equation}
where $u \in \{1, 2, \dots, K\}$, and $\text{FFNN}_\text{u}$ is a feedforward network mapping from $\mathbb{R}^{3 \times l} \rightarrow \mathbb{R}^p$.

Now, the most straightforward way to build the final pair representation $\textbf{f}_{ij}$ of $m_i$ and $m_j$ is to simply concatenate the trigger-based representation and all the feature-based representations together:
\begin{equation}\label{equa:simple_concat}
    \textbf{f}_{ij} = [\textbf{t}_{ij}, \textbf{h}_{ij}^{(1)}, \textbf{h}_{ij}^{(2)}, \dots, \textbf{h}_{ij}^{(K)}]
\end{equation}
However, this approach is not always optimal. First, as the symbolic features are predicted, they can be noisy and contain errors. The performance of most symbolic feature predictors is far from perfect (Table \ref{tab:attributes_list}). Also, depending on the specific context, some features can be more useful than others.

Inspired by studies on gated modules \cite{linetal2019reliability,gsamn}, we propose \textbf{Context-Dependent Gated Module} (CDGM), which uses a gating mechanism to extract information from the input symbolic features selectively (Figure \ref{fig:overall_pair_encoder}). Given two mentions $m_i$ and $m_j$, we use their trigger feature vector $\textbf{t}_{ij}$ as the main controlling context to compute the filtered representation $\overline{\textbf{h}}_{ij}^{\;(u)}$:
\begin{equation}
\overline{\textbf{h}}_{ij}^{\;(u)} = \text{CDGM}^{(u)}\big(\textbf{t}_{ij}, \textbf{h}_{ij}^{(u)}\big)
\end{equation}
where $u \in \{1, 2, \dots, K\}$. More specifically:
\begin{equation} \label{equa:cdgm}
\begin{split}
\textbf{g}^{(u)}_{ij} &= \sigma\big(\text{FFNN}_\text{g}^{(u)} \big(\big\lbrack \textbf{t}_{ij}, \textbf{h}_{ij}^{(u)}  \big\rbrack\big)\big) \\
\textbf{o}^{(u)}_{ij}, \textbf{p}^{(u)}_{ij} &= \text{DECOMPOSE}\big(\textbf{t}_{ij}, \textbf{h}_{ij}^{(u)}\big)\\
\overline{\textbf{h}}_{ij}^{\;(u)} &= \textbf{g}^{(u)}_{ij}\circ \textbf{o}^{(u)}_{ij} + \big(1 - \textbf{g}^{(u)}_{ij}\big)\circ\textbf{p}^{(u)}_{ij} 
\end{split}
\end{equation}
where $\sigma$ denotes sigmoid function. $\text{FFNN}_\text{g}^{(u)}$ is a mapping from $\mathbb{R}^{2 \times p} \rightarrow \mathbb{R}^p$. At a high level, $\textbf{h}_{ij}^{(u)}$ is decomposed into an orthogonal component and a parallel component, and $\overline{\textbf{h}}_{ij}^{\;(u)}$ is simply the fusion of these two components. In order to find the optimal mixture, $\textbf{g}_{ij}$ is used to control the composition. The decomposition unit is defined as:
\begin{equation} \label{equa:decomposition}
\begin{split}
&\text{\textit{Parallel}} \qquad\;\textbf{p}^{(u)}_{ij} = \frac{\textbf{h}_{ij}^{(u)} \,\cdot\,\textbf{t}_{ij}}{\textbf{t}_{ij}\,\cdot\,\textbf{t}_{ij}} \;\textbf{t}_{ij} \\
&\text{\textit{Orthogonal}} \quad \textbf{o}^{(u)}_{ij} = \textbf{h}_{ij}^{(u)} - \textbf{p}^{(u)}_{ij}\\
\end{split}
\end{equation}
where $\cdot$ denotes dot product. The parallel component $\textbf{p}_{ij}^{(u)}$ is the projection of $\textbf{h}_{ij}^{(u)}$ on $\textbf{t}_{ij}$. It can be viewed as containing information that is already part of $\textbf{t}_{ij}$. In contrast, $\textbf{o}_{ij}^{(u)}$ is orthogonal to $\textbf{t}_{ij}$, and so it can be viewed as containing \textit{new} information. Intuitively, when the original symbolic feature vector $\textbf{h}_{ij}^{(u)}$ is very clean and has complementary information, we want to utilize the new information in $\textbf{o}_{ij}^{(u)}$ (i.e., we want $\textbf{g}_{ij}^{(u)} \approx \textbf{1}$), and vice versa.

Finally, after using CDGMs to distill symbolic features, the final pair representation $\textbf{f}_{ij}$ of $m_i$ and $m_j$ can be computed as follows:
\begin{equation}
    \textbf{f}_{ij} = [\textbf{t}_{ij}, \overline{\textbf{h}}_{ij}^{\,(1)}, \overline{\textbf{h}}_{ij}^{\,(2)}, \dots, \overline{\textbf{h}}_{ij}^{\,(K)}]
\end{equation}
And the coreference score $s(i,j)$ of $m_i$ and $m_j$ is:
\begin{equation}
s(i, j) = \text{FFNN}_\text{a}(\textbf{f}_{ij})
\end{equation}
where $\text{FFNN}_\text{a}$ is a mapping from $\mathbb{R}^{(K+1) \times p} \rightarrow \mathbb{R}$.

% An advanced mechanism is clearly required to selectively extract information from the symbolic features.
\subsection{Training and Inference}
\begin{algorithm}[!ht]
\SetAlgoLined
\SetAlCapNameFnt{\small}
\SetAlCapFnt{\small}
\caption{Noise Addition for Symbolic Features}
\small
\textbf{Input:} Document $D$\\
\textbf{Hyperparameters}: $\{\epsilon_1, \epsilon_2, \cdots, \epsilon_K\}$\\
\For{$i = 1\;\dots\;k$}{
\For{$u = 1\;\dots\;K$}{
With prob. $\epsilon_u$, replace $c_{i}^{(u)}$ by $\hat{c}_{i}^{(u)} \sim \text{\textit{Uniform}}(N_u)$
}
}
\label{alg:noisy_training}
\end{algorithm}

\paragraph{Training} We use the same loss function as in \cite{Lee2017EndtoendNC}. Also, notice that the training accuracy of a feature predictor is typically much higher than its accuracy on the dev/test set (Table \ref{tab:attributes_list}). If we simply train our model without any regularization, our CDGMs will rarely come across noisy symbolic features during training. Therefore, to encourage our CDGMs to actually learn to distill reliable signals, we also propose a simple but effective \textbf{noisy training} method. Before passing a training data batch to the model, we randomly add noise to the predicted features. More specifically, for each document $D$ in the batch, we go through every symbolic feature of every event mention in $D$ and consider sampling a new value for the feature. The operation is described in Algorithm \ref{alg:noisy_training} (we use the same notations mentioned in Section \ref{sec:prelim}). $\{\epsilon_1, \epsilon_2, \cdots, \epsilon_K\}$ are hyperparameters determined by validation. In general, the larger the discrepancy between the train and test accuracies, the larger $\epsilon$.

\paragraph{Inference} For each (predicted) mention $m_i$, our model will assign an antecedent $a_i$ from all preceding mentions or a dummy antecedent $\epsilon$: $a_i \in Y(i) = \{\epsilon, m_1, m_2 \dots, m_{i-1}\}$. Basically, \(a_i = \argmax_{j < i} s(i, j)\). The dummy antecedent $\epsilon$ represents two possible cases: (1) $m_i$ is not actually an event mention (2) $m_i$ is indeed an event mention but it is not coreferent with any previous extracted mentions. In addition, we fix $s(i, \epsilon)$ to be 0.

\section{Experiments and Results}

% \heng{add (\%) into each result table caption. make the numbers consistent in each table, some are with \% and others are not. change 'performance' to F-score in caption}

%\subsection{Data and Experiments Setup}
\paragraph{Data and Experiments Setup} We evaluate our methods on two English datasets: ACE2005 \cite{walker2006ace} and KBP2016 \cite{ji2016overview,mitamura2016overview}. We report results in terms of F1 scores obtained using the CoNLL and AVG metrics. By definition, these metrics are the summary of other standard coreference metrics, including $\text{B}^3$, MUC, $\text{CEAF}_{e}$, and BLANC \cite{Lu2018EventCR}. We use SpanBERT (spanbert-base-cased) as the Transformer encoder \cite{Wolf2019HuggingFacesTS,Joshi2019SpanBERTIP}. More details about the datasets and hyperparameters are in the appendix. We refer to models that use only trigger features as [Baseline]. In a baseline model, $\textbf{f}_{ij}$ is simply $\textbf{t}_{ij}$ (Eq. \ref{equa:trigger_pair}). We refer to models that use only the simple concatenation strategy as [Simple] (Eq. \ref{equa:simple_concat}), and models that use the simple concatenation strategy and the noisy training method as [Noise].

%\subsection{Overall Performance}
%\subsection{Results and Analysis}

\renewcommand{\arraystretch}{1}
\begin{table}[!t]
\centering
\resizebox{0.85\linewidth}{!}{%
\begin{tabular}{lccccc}
\hline
ACE (Cross-Validation) & \multicolumn{1}{l}{CoNLL} & \multicolumn{1}{l}{AVG} \\ \hline
SSED + $\text{Supervised}_\text{Extended}$ \shortcite{pengetal2016event} & 55.23 & 52.53 \\
SSED + MSEP \shortcite{pengetal2016event} & 53.80 & 51.38 \\ \hline \\[-0.75em]
ACE (Test Data) & \multicolumn{1}{l}{CoNLL}  & \multicolumn{1}{l}{AVG} \\ \hline
Baseline & 58.93 & 55.78\\\hline
Simple (All Features) & 57.55 & 54.79 \\
CDGM (All Features) & 58.99 & 56.32 \\
Noise (All Features) & 60.43 & 57.85 \\
CDGM + Noise (All Features) & \textbf{62.07} & \textbf{59.76} \\ \hline
\end{tabular}%
}
\caption{End-to-end results on ACE 2005 (using predicted triggers and predicted symbolic features).}
\label{tab:end2end_overall_results_ace2005}
\end{table}

% \renewcommand{\arraystretch}{1}
% \begin{table*}[h!]
% \centering
% \resizebox{0.65\textwidth}{!}{%
% \begin{tabular}{lccccc}
% \hline
% ACE (Cross-Validation) & \multicolumn{1}{l}{$B^3$} & \multicolumn{1}{l}{MUC} & \multicolumn{1}{l}{$\text{CEAF}_{e}$} & \multicolumn{1}{l}{BLANC} & \multicolumn{1}{l}{AVG} \\ \hline
% SSED + $\text{Supervised}_\text{Extended}$ \shortcite{pengetal2016event} & 59.90 & 47.10 & 58.70 & 44.40 & 52.53 \\
% SSED + MSEP \shortcite{pengetal2016event} & 60.30 & 42.10 & 59.00 & 44.10 & 51.38 \\ \hline \\[-0.75em]
% ACE (Test Data) & \multicolumn{1}{l}{$B^3$} & \multicolumn{1}{l}{MUC} & \multicolumn{1}{l}{$\text{CEAF}_{e}$} & \multicolumn{1}{l}{BLANC} & \multicolumn{1}{l}{AVG} \\ \hline
% Baseline & 64.31 & 50.00 & 62.47 & 46.32 & 55.78\\\hline
% Simple (All Features) & 65.50 & 43.92 & 63.24 & 46.48 & 54.79 \\
% CDGM (All Features) & 65.94 & 47.22 & 63.80 & 48.32 & 56.32 \\
% Noise (All Features) & 66.76 & 51.12 & 63.40 & 50.10 & 57.85 \\
% CDGM + Noise (All Features) & \textbf{67.18} & \textbf{54.00} & \textbf{65.03} & \textbf{52.81} & \textbf{59.76} \\ \hline
% \end{tabular}%
% }
% \caption{Overall performance on the ACE 2005 dataset.}
% \label{tab:end2end_overall_results_ace2005}
% \end{table*}

% For our systems, event mentions and their types are extracted by OneIE \cite{linetal2020joint}, while other attributes are predicted by a simple Transformer-based model.

\renewcommand{\arraystretch}{1}
\begin{table}[!t]
\centering
\resizebox{0.85\linewidth}{!}{%
\begin{tabular}{lccccc}
\hline
System & \multicolumn{1}{l}{CoNLL} & \multicolumn{1}{l}{AVG} \\ \hline
UTD's system \shortcite{Lu2015UTDsEN}  & 32.69 & 30.08 \\
%TAMU's system \shortcite{Choubey2017TAMUAK} & 29.91 & 27.13 \\ 
Joint Learning \shortcite{lung2017joint} & 35.77 & 33.08 \\
$\text{E}^3\text{C}$ \shortcite{Lu2020EndtoEndNE} & 41.97 & 38.66  \\\hline
Baseline & 40.57 & 37.59 \\\hline
Simple (All Features) & 41.40 & 38.58 \\
CDGM + Noise (All Features) & \textbf{43.55} & \textbf{40.61} \\\hline
\end{tabular}%
}
\caption{End-to-end results on KBP 2016 (using predicted triggers and predicted symbolic features).}
\label{tab:end2end_overall_results_kbp2016}
%\vspace{-4mm}
\end{table}

% \renewcommand{\arraystretch}{1}
% \begin{table*}[!h]
% \centering
% \resizebox{0.65\linewidth}{!}{%
% \begin{tabular}{lccccc}
% \hline
% System & \multicolumn{1}{l}{$B^3$} & \multicolumn{1}{l}{MUC} & \multicolumn{1}{l}{$\text{CEAF}_{e}$} & \multicolumn{1}{l}{BLANC} & \multicolumn{1}{l}{AVG} \\ \hline
% UTD's system \shortcite{Lu2015UTDsEN} & 37.49 & 26.37 & 34.21 & 22.25 & 30.08 \\
% TAMU's system \shortcite{Choubey2017TAMUAK} & 36.62 & 17.62 & 35.50 & 18.77 & 27.13 \\ 
% Joint Learning \shortcite{lung2017joint} & 40.90 & 27.41 & 39.00 & 25.00 & 33.08 \\
% $\text{E}^3\text{C}$ \shortcite{Lu2020EndtoEndNE} & 46.32 & \textbf{34.39} & 45.19 & 28.74 & 38.66  \\\hline
% Baseline & 46.65 & 28.34 & 46.73 & 28.62 & 37.59 \\\hline
% Simple (All Features) & 47.73 & 27.42 & 49.05 & 30.11 & 38.58 \\
% CDGM + Noise (All Features) & \textbf{48.98} & 30.78 & \textbf{50.88} & \textbf{31.78} & \textbf{40.61} \\\hline
% \end{tabular}%
% }
% \caption{Overall performance on the KBP 2016 dataset.}
% \label{tab:end2end_overall_results_kbp2016}
% \end{table*}

\paragraph{Overall Results (on Predicted Mentions)} Table \ref{tab:end2end_overall_results_ace2005} and Table \ref{tab:end2end_overall_results_kbp2016} show the overall end-to-end results on ACE2005 and KBP2016, respectively. We use OneIE \cite{linetal2020joint} to extract event mentions and their types. Other features are predicted by a simple Transformer model. Overall, our full model outperforms the baseline model by a large margin and significantly outperforms state-of-the-art on KBP 2016. Our ACE 2005 scores are not directly comparable with previous work, as \newcite{pengetal2016event} conducted 10-fold cross-validation and essentially used more training data. Nevertheless, the magnitude of the differences in scores between our best model and the state-of-the-art methods indicates the effectiveness of our methods.

\renewcommand{\arraystretch}{1}
\begin{table}[!t]
\centering
\resizebox{0.85\linewidth}{!}{%
\begin{tabular}{lccccc}
\hline
ACE (Test Data) & \multicolumn{1}{l}{CoNLL} & \multicolumn{1}{l}{AVG} \\ \hline
PAIREDRL \shortcite{Yu2020PairedRL} & 84.65 & - \\\hline
Baseline & 81.62 & 81.49 \\
Simple (All Features) & 75.32 & 74.94\\
CDGM + Noise (All Features) & \textbf{84.76} & \textbf{83.95} \\ \hline
\end{tabular}%
}
\caption{Results on ACE 2005 using gold triggers and predicted symbolic features.}
\label{tab:gold_trigger_overall_results_ace2005}
%\vspace{-3mm}
\end{table}

% ground-truth
\renewcommand{\arraystretch}{1}
\begin{table}[!t]
\centering
\resizebox{0.85\linewidth}{!}{%
\begin{tabular}{lccccc}
\hline
ACE (Test Data) & \multicolumn{1}{l}{CoNLL} & \multicolumn{1}{l}{AVG} \\ \hline
Baseline & 81.62 & 81.49 \\
Simple (All Features) & 85.75 & 85.40 \\
CDGM (All Features) & \textbf{87.90} & \textbf{88.30} \\
CDGM + Noise (All Features) & 85.40 & 85.38 \\\hline
\end{tabular}%
}
\caption{Results on ACE 2005 using gold triggers and ground-truth symbolic features.}
\label{tab:gold_trigger_gold_attributes_ace2005}
%\vspace{-3mm}
\end{table}

% ground-truth

\paragraph{Overall Results (on Ground-truth Triggers)} The overall results on ACE 2005 using ground-truth triggers and predicted symbolic features are shown in Table \ref{tab:gold_trigger_overall_results_ace2005}. The performance of our full model is comparable with previous state-of-the-art result in \cite{Yu2020PairedRL}. To better analyze the usefulness of symbolic features as well as the effectiveness of our methods, we also conduct experiments using \textit{ground-truth} triggers and \textit{ground-truth} symbolic features (Table \ref{tab:gold_trigger_gold_attributes_ace2005}). First, when the symbolic features are clean, incorporating them using the simple concatenation strategy can already boost the performance significantly. The symbolic features contain information complementary to that in the SpanBERT contextual embeddings. Second, we also see that the noisy training method is not helpful when the symbolic features are clean. Unlike other regularization methods such as dropout \cite{dropout} and weight decay \cite{NIPS1991_8eefcfdf}, the main role of our noisy training method is not to reduce overfitting in the traditional sense. Its main function is to help CDGMs learn to distill reliable signals from noisy features.
%\subsection{Results of Incorporating Different Types of Symbolic Features}
\paragraph{Impact of Different Symbolic Features} Table \ref{tab:ace_individual_features} shows the results of incorporating different types of symbolic features on the ACE 2005 dataset. Overall, our methods consistently perform better than the simple concatenation strategy across all feature types. The gains are also larger for more noisy features than clean features (feature prediction accuracies were shown in Table \ref{tab:attributes_list}). This suggests that our methods are particularly useful in situations where the symbolic features are noisy.

\begin{table}[!t]
\centering
\resizebox{0.85\linewidth}{!}{%
\begin{tabular}{lccc}
\hline
Features & \begin{tabular}[c]{@{}c@{}}AVG\\ (Simple)\end{tabular} & \begin{tabular}[c]{@{}c@{}}AVG\\ (CDGM + Noise)\end{tabular} & $\Delta_\text{AVG}$ \\ \hline
Subtype & 56.41 & 57.02 & \textbf{+0.61} \\ 
Polarity & 56.06 & 57.03 & \textbf{+0.97} \\
Modality & 54.81 & 58.54 & \textbf{+3.73} \\
Genericity & 54.70 & 57.82 & \textbf{+3.12}\\
Tense & 54.28 & 56.62 &  \textbf{+2.34}\\\hline
\end{tabular}%
}
%\caption{Results of incorporating various types of symbolic features (ACE 2005 dataset).}
\caption{Impact of Symbolic Features (ACE 2005)}
%Results of incorporating various types of symbolic features (ACE 2005 dataset).}
\label{tab:ace_individual_features}
\vspace{-4mm}
\end{table}
\paragraph{Comparison with Multi-Task Learning} We also investigate whether we can incorporate symbolic semantics into coreference resolution by simply doing multi-task training. We train our baseline model to jointly perform coreference resolution and symbolic feature prediction. The test AVG score on ACE 2005 is only 56.5. In contrast, our best model achieves an AVG score of 59.76 (Table \ref{tab:end2end_overall_results_ace2005}).
\paragraph{Qualitative Examples} Table \ref{tab:qualitative_examples} shows few examples from the ACE 2005 dataset that illustrate how incorporating symbolic features using our proposed methods can improve the performance of event conference resolution. In each example, our baseline model incorrectly predicts the highlighted event mentions to be coreferential.

\paragraph{Remaining Challenges} Previous studies suggest that there exist different types and degrees of event coreference \cite{Recasens2011IdentityNA,hovyetal2013events}. Many methods (including ours) focus on the full strict coreference task, but other types of coreference such as partial coreference have remained underexplored. \newcite{hovyetal2013events} defines two core types of partial event coreference relations: subevent relations and membership relations. Subevent relations form a stereotypical sequence of events, whereas membership relations represent instances of an event collection. We leave tackling the partial coreference task to future work.

\renewcommand{\arraystretch}{1.25}
\begin{table}[t!]
\centering
\small
\begin{tabular}{p{0.92\linewidth}}
\hline
... \textcolor{red}{$\{$Negotiations$\}_{\text{ev1}}$} between Washington and ... \\
... think that this will affect the \textcolor{blue}{$\{$elections$\}_{\text{ev2}}$} unless ...\\
\textcolor{red}{\text{ev1}} (\texttt{Contact:Meet})\\
\textcolor{blue}{\text{ev2}} (\texttt{Personnel:Elect})\\
\hline
... since you are \textbf{not} directly \textcolor{red}{$\{$elected$\}_{\text{ev1}}$}, it would be ... \\
... Az-Zaman daily that \textcolor{blue}{$\{$elections$\}_{\text{ev2}}$} should be held ... \\
\textcolor{red}{\text{ev1}} (\texttt{Personnel:Elect}): Polarity = NEGATIVE\\
\textcolor{blue}{\text{ev2}} (\texttt{Personnel:Elect}): Polarity = POSITIVE\\
\hline
... told reporters after his \textcolor{red}{$\{$appeal$\}_{\text{ev1}}$} was rejected ...\\
... most junior of the court of \textcolor{blue}{$\{$appeal$\}_{\text{ev2}}$}, and its ...\\
\textcolor{red}{\text{ev1}} (\texttt{Justice:Appeal}): Genericity = SPECIFIC\\
\textcolor{blue}{\text{ev2}} (\texttt{Justice:Appeal}): Genericity = GENERIC\\
\hline
\end{tabular}
\caption{\label{tab:qualitative_examples} Examples of using symbolic features to improve event coreference resolution.}
\vspace{-6mm}
\end{table}
\section{Related Work}

Several previous approaches to within-document event coreference resolution operate by first applying a mention-pair model to compute pairwise distances between event mentions, and then they apply a clustering algorithm such as agglomerative clustering or spectral graph clustering \cite{chenetal2009pairwise,chenji2009graph,Chen2014SinoCoreferencerAE,nguyen2016new,huangetal2019improving}. In addition to trigger features, these methods use a variety of additional symbolic features such as event types, attributes, arguments, and distance. These approaches do not use contextual embeddings such as BERT and SpanBERT \cite{devlinetal2019bert,Joshi2019SpanBERTIP}. Recently, there are several studies that use contextual embeddings together with type-based or argument-based information \cite{Lu2020EndtoEndNE,Yu2020PairedRL}. These methods design networks or mechanisms that are specific to only one type of symbolic features. In contrast, our work is more general and can be effectively applied to a wide range of symbolic features.

\section{Conclusions and Future Work}
In this work, we propose a novel gated module to incorporate symbolic semantics into event coreference resolution. Combined with a simple noisy training technique, our best models achieve competitive results on ACE 2005 and KBP 2016. In the future, we aim to extend our work to address more general problems such as cross-lingual cross-document coreference resolution.

\section*{Acknowledgement}
This research is based upon work supported in part by U.S. DARPA KAIROS Program No. FA8750-19-2-1004, U.S. DARPA AIDA Program No. FA8750-18-2-0014, and Air Force No. FA8650-17-C-7715. The views and conclusions contained herein are those of the authors and should not be interpreted as necessarily representing the official policies, either expressed or implied, of DARPA, or the U.S. Government. The U.S. Government is authorized to reproduce and distribute reprints for governmental purposes notwithstanding any copyright annotation therein.

\bibliography{anthology,naacl2021}

\begin{thebibliography}{42}
\expandafter\ifx\csname natexlab\endcsname\relax\def\natexlab#1{#1}\fi

\bibitem[{Chen and Ng(2014)}]{Chen2014SinoCoreferencerAE}
C.~Chen and Vincent Ng. 2014.
\newblock Sinocoreferencer: An end-to-end chinese event coreference resolver.
\newblock In \emph{LREC}.

\bibitem[{Chen and Ng(2016)}]{Chen2016JointIO}
Chen Chen and Vincent Ng. 2016.
\newblock Joint inference over a lightly supervised information extraction
  pipeline: Towards event coreference resolution for resource-scarce languages.
\newblock In \emph{Proceedings of the Thirtieth AAAI Conference on Artificial
  Intelligence}, AAAI'16, page 2913–2920. AAAI Press.

\bibitem[{Chen et~al.(2015)Chen, Xu, Liu, Zeng, and Zhao}]{chenetal2015event}
Yubo Chen, Liheng Xu, Kang Liu, Daojian Zeng, and Jun Zhao. 2015.
\newblock \href {https://doi.org/10.3115/v1/P15-1017} {Event extraction via
  dynamic multi-pooling convolutional neural networks}.
\newblock In \emph{Proceedings of the 53rd Annual Meeting of the Association
  for Computational Linguistics and the 7th International Joint Conference on
  Natural Language Processing (Volume 1: Long Papers)}, pages 167--176,
  Beijing, China. Association for Computational Linguistics.

\bibitem[{Chen and Ji(2009)}]{chenji2009graph}
Zheng Chen and Heng Ji. 2009.
\newblock \href {https://www.aclweb.org/anthology/W09-3208} {Graph-based event
  coreference resolution}.
\newblock In \emph{Proceedings of the 2009 Workshop on Graph-based Methods for
  Natural Language Processing ({T}ext{G}raphs-4)}, pages 54--57, Suntec,
  Singapore. Association for Computational Linguistics.

\bibitem[{Chen et~al.(2009)Chen, Ji, and Haralick}]{chenetal2009pairwise}
Zheng Chen, Heng Ji, and Robert Haralick. 2009.
\newblock \href {https://www.aclweb.org/anthology/W09-4303} {A pairwise event
  coreference model, feature impact and evaluation for event coreference
  resolution}.
\newblock In \emph{Proceedings of the Workshop on Events in Emerging Text
  Types}, pages 17--22, Borovets, Bulgaria. Association for Computational
  Linguistics.

\bibitem[{Choubey and Huang(2017)}]{Choubey2017EventCR}
Prafulla~Kumar Choubey and Ruihong Huang. 2017.
\newblock \href {https://doi.org/10.18653/v1/D17-1226} {Event coreference
  resolution by iteratively unfolding inter-dependencies among events}.
\newblock In \emph{Proceedings of the 2017 Conference on Empirical Methods in
  Natural Language Processing}, pages 2124--2133, Copenhagen, Denmark.
  Association for Computational Linguistics.

\bibitem[{Choubey et~al.(2018)Choubey, Raju, and
  Huang}]{Choubey2018IdentifyingTM}
Prafulla~Kumar Choubey, Kaushik Raju, and Ruihong Huang. 2018.
\newblock \href {https://doi.org/10.18653/v1/N18-2055} {Identifying the most
  dominant event in a news article by mining event coreference relations}.
\newblock In \emph{Proceedings of the 2018 Conference of the North {A}merican
  Chapter of the Association for Computational Linguistics: Human Language
  Technologies, Volume 2 (Short Papers)}, pages 340--345, New Orleans,
  Louisiana. Association for Computational Linguistics.

\bibitem[{Devlin et~al.(2019)Devlin, Chang, Lee, and
  Toutanova}]{devlinetal2019bert}
Jacob Devlin, Ming-Wei Chang, Kenton Lee, and Kristina Toutanova. 2019.
\newblock \href {https://doi.org/10.18653/v1/N19-1423} {{BERT}: Pre-training of
  deep bidirectional transformers for language understanding}.
\newblock In \emph{Proceedings of the 2019 Conference of the North {A}merican
  Chapter of the Association for Computational Linguistics: Human Language
  Technologies, Volume 1 (Long and Short Papers)}, pages 4171--4186,
  Minneapolis, Minnesota. Association for Computational Linguistics.

\bibitem[{Duncan et~al.(2017)Duncan, Chan, Peng, Wu, Upadhyay, Gupta, Tsai,
  Sammons, and Roth}]{duncan2017ui}
Chase Duncan, Liang{-}Wei Chan, Haoruo Peng, Hao Wu, Shyam Upadhyay, Nitish
  Gupta, Chen{-}Tse Tsai, Mark Sammons, and Dan Roth. 2017.
\newblock \href
  {https://tac.nist.gov/publications/2017/participant.papers/TAC2017.UI\_CCG.proceedings.pdf}
  {{UI} {CCG} {TAC-KBP2017} submissions: Entity discovery and linking, and
  event nugget detection and co-reference}.
\newblock In \emph{Proceedings of the 2017 Text Analysis Conference, {TAC}
  2017, Gaithersburg, Maryland, USA, November 13-14, 2017}. {NIST}.

\bibitem[{Harris et~al.(2020)Harris, Millman, Walt, Gommers, Virtanen,
  Cournapeau, Wieser, Taylor, Berg, Smith, Kern, Picus, Hoyer, Kerkwijk, Brett,
  Haldane, del R'io, Wiebe, Peterson, G'erard-Marchant, Sheppard, Reddy,
  Weckesser, Abbasi, Gohlke, and Oliphant}]{Harris2020ArrayPW}
C.~Harris, K.~J. Millman, S.~Walt, Ralf Gommers, P.~Virtanen, D.~Cournapeau,
  E.~Wieser, J.~Taylor, S.~Berg, Nathaniel~J. Smith, R.~Kern, Matti Picus,
  S.~Hoyer, M.~Kerkwijk, Matthew Brett, Allan Haldane, Jaime~Fern'andez del
  R'io, Mark Wiebe, P.~Peterson, Pierre G'erard-Marchant, K.~Sheppard,
  T.~Reddy, W.~Weckesser, H.~Abbasi, Christoph Gohlke, and T.~E. Oliphant.
  2020.
\newblock Array programming with numpy.
\newblock \emph{Nature}, 585 7825:357--362.

\bibitem[{Hovy et~al.(2013)Hovy, Mitamura, Verdejo, Araki, and
  Philpot}]{hovyetal2013events}
Eduard Hovy, Teruko Mitamura, Felisa Verdejo, Jun Araki, and Andrew Philpot.
  2013.
\newblock \href {https://www.aclweb.org/anthology/W13-1203} {Events are not
  simple: Identity, non-identity, and quasi-identity}.
\newblock In \emph{Workshop on Events: Definition, Detection, Coreference, and
  Representation}, pages 21--28, Atlanta, Georgia. Association for
  Computational Linguistics.

\bibitem[{Huang et~al.(2019)Huang, Lu, Kurohashi, and
  Ng}]{huangetal2019improving}
Yin~Jou Huang, Jing Lu, Sadao Kurohashi, and Vincent Ng. 2019.
\newblock \href {https://doi.org/10.18653/v1/N19-1085} {Improving event
  coreference resolution by learning argument compatibility from unlabeled
  data}.
\newblock In \emph{Proceedings of the 2019 Conference of the North {A}merican
  Chapter of the Association for Computational Linguistics: Human Language
  Technologies, Volume 1 (Long and Short Papers)}, pages 785--795, Minneapolis,
  Minnesota. Association for Computational Linguistics.

\bibitem[{Ji and Grishman(2011)}]{jigrishman2011knowledge}
Heng Ji and Ralph Grishman. 2011.
\newblock \href {https://www.aclweb.org/anthology/P11-1115} {Knowledge base
  population: Successful approaches and challenges}.
\newblock In \emph{Proceedings of the 49th Annual Meeting of the Association
  for Computational Linguistics: Human Language Technologies}, pages
  1148--1158, Portland, Oregon, USA. Association for Computational Linguistics.

\bibitem[{Ji et~al.(2016)Ji, Nothman, Dang, and Hub}]{ji2016overview}
Heng Ji, Joel Nothman, H~Trang Dang, and Sydney~Informatics Hub. 2016.
\newblock Overview of tac-kbp2016 tri-lingual edl and its impact on end-to-end
  cold-start kbp.
\newblock \emph{Proceedings of TAC}.

\bibitem[{Joshi et~al.(2020)Joshi, Chen, Liu, Weld, Zettlemoyer, and
  Levy}]{Joshi2019SpanBERTIP}
Mandar Joshi, Danqi Chen, Yinhan Liu, Daniel Weld, Luke Zettlemoyer, and Omer
  Levy. 2020.
\newblock \href {https://transacl.org/ojs/index.php/tacl/article/view/1853}
  {Spanbert: Improving pre-training by representing and predicting spans}.
\newblock \emph{Transactions of the Association for Computational Linguistics},
  8(0):64--77.

\bibitem[{Krogh and Hertz(1992)}]{NIPS1991_8eefcfdf}
Anders Krogh and John Hertz. 1992.
\newblock \href
  {https://proceedings.neurips.cc/paper/1991/file/8eefcfdf5990e441f0fb6f3fad709e21-Paper.pdf}
  {A simple weight decay can improve generalization}.
\newblock In \emph{Advances in Neural Information Processing Systems},
  volume~4. Morgan-Kaufmann.

\bibitem[{Lai et~al.(2019)Lai, Tran, Bui, and Kihara}]{gsamn}
Tuan Lai, Quan~Hung Tran, Trung Bui, and Daisuke Kihara. 2019.
\newblock \href {https://doi.org/10.18653/v1/D19-1610} {A gated self-attention
  memory network for answer selection}.
\newblock In \emph{Proceedings of the 2019 Conference on Empirical Methods in
  Natural Language Processing and the 9th International Joint Conference on
  Natural Language Processing (EMNLP-IJCNLP)}, pages 5953--5959, Hong Kong,
  China. Association for Computational Linguistics.

\bibitem[{LDC(2005)}]{ace2005english}
LDC. 2005.
\newblock English annotation guidelines for events version 5.4. 3.
\newblock \emph{Linguistic Data Consortium}, 1.

\bibitem[{Lee et~al.(2017)Lee, He, Lewis, and Zettlemoyer}]{Lee2017EndtoendNC}
Kenton Lee, Luheng He, Mike Lewis, and Luke Zettlemoyer. 2017.
\newblock \href {https://doi.org/10.18653/v1/D17-1018} {End-to-end neural
  coreference resolution}.
\newblock In \emph{Proceedings of the 2017 Conference on Empirical Methods in
  Natural Language Processing}, pages 188--197, Copenhagen, Denmark.
  Association for Computational Linguistics.

\bibitem[{Lin et~al.(2020)Lin, Ji, Huang, and Wu}]{linetal2020joint}
Ying Lin, Heng Ji, Fei Huang, and Lingfei Wu. 2020.
\newblock \href {https://doi.org/10.18653/v1/2020.acl-main.713} {A joint neural
  model for information extraction with global features}.
\newblock In \emph{Proceedings of the 58th Annual Meeting of the Association
  for Computational Linguistics}, pages 7999--8009, Online. Association for
  Computational Linguistics.

\bibitem[{Lin et~al.(2019)Lin, Liu, Ji, Yu, and Han}]{linetal2019reliability}
Ying Lin, Liyuan Liu, Heng Ji, Dong Yu, and Jiawei Han. 2019.
\newblock \href {https://doi.org/10.18653/v1/P19-1016} {Reliability-aware
  dynamic feature composition for name tagging}.
\newblock In \emph{Proceedings of the 57th Annual Meeting of the Association
  for Computational Linguistics}, pages 165--174, Florence, Italy. Association
  for Computational Linguistics.

\bibitem[{Lu and Ng(2017{\natexlab{a}})}]{Lu2017LearningAS}
J.~Lu and Vincent Ng. 2017{\natexlab{a}}.
\newblock Learning antecedent structures for event coreference resolution.
\newblock \emph{2017 16th IEEE International Conference on Machine Learning and
  Applications (ICMLA)}, pages 113--118.

\bibitem[{Lu and Ng(2015)}]{Lu2015UTDsEN}
Jing Lu and Vincent Ng. 2015.
\newblock \href
  {https://tac.nist.gov/publications/2015/participant.papers/TAC2015.UTD.proceedings.pdf}
  {Utd's event nugget detection and coreference system at {KBP} 2015}.
\newblock In \emph{Proceedings of the 2015 Text Analysis Conference, {TAC}
  2015, Gaithersburg, Maryland, USA, November 16-17, 2015, 2015}. {NIST}.

\bibitem[{Lu and Ng(2016)}]{Lu2016EventCR}
Jing Lu and Vincent Ng. 2016.
\newblock \href {https://www.aclweb.org/anthology/L16-1631} {Event coreference
  resolution with multi-pass sieves}.
\newblock In \emph{Proceedings of the Tenth International Conference on
  Language Resources and Evaluation ({LREC}'16)}, pages 3996--4003,
  Portoro{\v{z}}, Slovenia. European Language Resources Association (ELRA).

\bibitem[{Lu and Ng(2017{\natexlab{b}})}]{lung2017joint}
Jing Lu and Vincent Ng. 2017{\natexlab{b}}.
\newblock \href {https://doi.org/10.18653/v1/P17-1009} {Joint learning for
  event coreference resolution}.
\newblock In \emph{Proceedings of the 55th Annual Meeting of the Association
  for Computational Linguistics (Volume 1: Long Papers)}, pages 90--101,
  Vancouver, Canada. Association for Computational Linguistics.

\bibitem[{Lu and Ng(2018)}]{Lu2018EventCR}
Jing Lu and Vincent Ng. 2018.
\newblock \href {https://doi.org/10.24963/ijcai.2018/773} {Event coreference
  resolution: A survey of two decades of research}.
\newblock In \emph{Proceedings of the Twenty-Seventh International Joint
  Conference on Artificial Intelligence, {IJCAI-18}}, pages 5479--5486.
  International Joint Conferences on Artificial Intelligence Organization.

\bibitem[{Lu et~al.(2020)Lu, Lin, Tang, Han, and Sun}]{Lu2020EndtoEndNE}
Yaojie Lu, Hongyu Lin, Jialong Tang, Xianpei Han, and Le~Sun. 2020.
\newblock End-to-end neural event coreference resolution.
\newblock \emph{arXiv preprint arXiv:2009.08153}.

\bibitem[{Mikolov et~al.(2013)Mikolov, Sutskever, Chen, Corrado, and
  Dean}]{word2vec}
Tomas Mikolov, Ilya Sutskever, Kai Chen, Greg Corrado, and Jeffrey Dean. 2013.
\newblock Distributed representations of words and phrases and their
  compositionality.
\newblock In \emph{Proceedings of the 26th International Conference on Neural
  Information Processing Systems - Volume 2}, NIPS'13, page 3111–3119, Red
  Hook, NY, USA. Curran Associates Inc.

\bibitem[{Mitamura et~al.(2016)Mitamura, Liu, and Hovy}]{mitamura2016overview}
Teruko Mitamura, Zhengzhong Liu, and Eduard~H. Hovy. 2016.
\newblock \href
  {https://tac.nist.gov/publications/2016/additional.papers/TAC2016.KBP\_Event\_Nugget\_overview.proceedings.pdf}
  {Overview of {TAC-KBP} 2016 event nugget track}.
\newblock In \emph{Proceedings of the 2016 Text Analysis Conference, {TAC}
  2016, Gaithersburg, Maryland, USA, November 14-15, 2016}. {NIST}.

\bibitem[{Nguyen et~al.(2016)Nguyen, Meyers, and Grishman}]{nguyen2016new}
Thien~Huu Nguyen, Adam Meyers, and Ralph Grishman. 2016.
\newblock \href
  {https://tac.nist.gov/publications/2016/participant.papers/TAC2016.NYU.proceedings.pdf}
  {New york university 2016 system for {KBP} event nugget: {A} deep learning
  approach}.
\newblock In \emph{Proceedings of the 2016 Text Analysis Conference, {TAC}
  2016, Gaithersburg, Maryland, USA, November 14-15, 2016}. {NIST}.

\bibitem[{Paszke et~al.(2019)Paszke, Gross, Massa, Lerer, Bradbury, Chanan,
  Killeen, Lin, Gimelshein, Antiga, Desmaison, K{\"o}pf, Yang, DeVito, Raison,
  Tejani, Chilamkurthy, Steiner, Fang, Bai, and Chintala}]{Paszke2019PyTorchAI}
Adam Paszke, S.~Gross, Francisco Massa, A.~Lerer, J.~Bradbury, G.~Chanan,
  T.~Killeen, Z.~Lin, N.~Gimelshein, L.~Antiga, Alban Desmaison, Andreas
  K{\"o}pf, E.~Yang, Zach DeVito, Martin Raison, Alykhan Tejani, Sasank
  Chilamkurthy, B.~Steiner, Lu~Fang, Junjie Bai, and Soumith Chintala. 2019.
\newblock Pytorch: An imperative style, high-performance deep learning library.
\newblock \emph{ArXiv}, abs/1912.01703.

\bibitem[{Peng et~al.(2016)Peng, Song, and Roth}]{pengetal2016event}
Haoruo Peng, Yangqiu Song, and Dan Roth. 2016.
\newblock \href {https://doi.org/10.18653/v1/D16-1038} {Event detection and
  co-reference with minimal supervision}.
\newblock In \emph{Proceedings of the 2016 Conference on Empirical Methods in
  Natural Language Processing}, pages 392--402, Austin, Texas. Association for
  Computational Linguistics.

\bibitem[{Pennington et~al.(2014)Pennington, Socher, and
  Manning}]{penningtonetal2014glove}
Jeffrey Pennington, Richard Socher, and Christopher Manning. 2014.
\newblock \href {https://doi.org/10.3115/v1/D14-1162} {{G}lo{V}e: Global
  vectors for word representation}.
\newblock In \emph{Proceedings of the 2014 Conference on Empirical Methods in
  Natural Language Processing ({EMNLP})}, pages 1532--1543, Doha, Qatar.
  Association for Computational Linguistics.

\bibitem[{Recasens et~al.(2011)Recasens, Hovy, and
  Mart{\'i}}]{Recasens2011IdentityNA}
M.~Recasens, E.~Hovy, and M.~Mart{\'i}. 2011.
\newblock Identity, non-identity, and near-identity: Addressing the complexity
  of coreference.
\newblock \emph{Lingua}, 121:1138--1152.

\bibitem[{Sammons et~al.(2015)Sammons, Peng, Song, Upadhyay, Tsai, Reddy, Roy,
  and Roth}]{sammons2015illinois}
Mark Sammons, Haoruo Peng, Yangqiu Song, Shyam Upadhyay, Chen{-}Tse Tsai,
  Pavankumar Reddy, Subhro Roy, and Dan Roth. 2015.
\newblock \href
  {https://tac.nist.gov/publications/2015/participant.papers/TAC2015.UI\_CCG.proceedings.pdf}
  {Illinois {CCG} {TAC} 2015 event nugget, entity discovery and linking, and
  slot filler validation systems}.
\newblock In \emph{Proceedings of the 2015 Text Analysis Conference, {TAC}
  2015, Gaithersburg, Maryland, USA, November 16-17, 2015, 2015}. {NIST}.

\bibitem[{Srivastava et~al.(2014)Srivastava, Hinton, Krizhevsky, Sutskever, and
  Salakhutdinov}]{dropout}
Nitish Srivastava, Geoffrey Hinton, Alex Krizhevsky, Ilya Sutskever, and Ruslan
  Salakhutdinov. 2014.
\newblock \href {http://jmlr.org/papers/v15/srivastava14a.html} {Dropout: A
  simple way to prevent neural networks from overfitting}.
\newblock \emph{Journal of Machine Learning Research}, 15(56):1929--1958.

\bibitem[{Vanderwende et~al.(2004)Vanderwende, Banko, and
  Menezes}]{Vanderwende2004EventCentricSG}
Lucy Vanderwende, Michele Banko, and Arul Menezes. 2004.
\newblock \href
  {https://www.microsoft.com/en-us/research/publication/event-centric-summary-generation/}
  {Event-centric summary generation}.
\newblock In \emph{Working notes of the Document Understanding Conference
  2004}. ACL.

\bibitem[{Walker et~al.(2006)Walker, Strassel, Medero, and
  Maeda}]{walker2006ace}
Christopher Walker, Stephanie Strassel, Julie Medero, and Kazuaki Maeda. 2006.
\newblock Ace 2005 multilingual training corpus.
\newblock \emph{Linguistic Data Consortium, Philadelphia}, 57:45.

\bibitem[{Wolf et~al.(2020{\natexlab{a}})Wolf, Debut, Sanh, Chaumond, Delangue,
  Moi, Cistac, Rault, Louf, Funtowicz, Davison, Shleifer, von Platen, Ma,
  Jernite, Plu, Xu, Le~Scao, Gugger, Drame, Lhoest, and
  Rush}]{Wolf2019HuggingFacesTS}
Thomas Wolf, Lysandre Debut, Victor Sanh, Julien Chaumond, Clement Delangue,
  Anthony Moi, Pierric Cistac, Tim Rault, Remi Louf, Morgan Funtowicz, Joe
  Davison, Sam Shleifer, Patrick von Platen, Clara Ma, Yacine Jernite, Julien
  Plu, Canwen Xu, Teven Le~Scao, Sylvain Gugger, Mariama Drame, Quentin Lhoest,
  and Alexander Rush. 2020{\natexlab{a}}.
\newblock \href {https://www.aclweb.org/anthology/2020.emnlp-demos.6}
  {Transformers: State-of-the-art natural language processing}.
\newblock In \emph{Proceedings of the 2020 Conference on Empirical Methods in
  Natural Language Processing: System Demonstrations}, pages 38--45, Online.
  Association for Computational Linguistics.

\bibitem[{Wolf et~al.(2020{\natexlab{b}})Wolf, Debut, Sanh, Chaumond, Delangue,
  Moi, Cistac, Rault, Louf, Funtowicz, Davison, Shleifer, von Platen, Ma,
  Jernite, Plu, Xu, Le~Scao, Gugger, Drame, Lhoest, and
  Rush}]{wolfetal2020transformers}
Thomas Wolf, Lysandre Debut, Victor Sanh, Julien Chaumond, Clement Delangue,
  Anthony Moi, Pierric Cistac, Tim Rault, Remi Louf, Morgan Funtowicz, Joe
  Davison, Sam Shleifer, Patrick von Platen, Clara Ma, Yacine Jernite, Julien
  Plu, Canwen Xu, Teven Le~Scao, Sylvain Gugger, Mariama Drame, Quentin Lhoest,
  and Alexander Rush. 2020{\natexlab{b}}.
\newblock \href {https://www.aclweb.org/anthology/2020.emnlp-demos.6}
  {Transformers: State-of-the-art natural language processing}.
\newblock In \emph{Proceedings of the 2020 Conference on Empirical Methods in
  Natural Language Processing: System Demonstrations}, pages 38--45, Online.
  Association for Computational Linguistics.

\bibitem[{Yu et~al.(2020)Yu, Yin, and Roth}]{Yu2020PairedRL}
Xiaodong Yu, Wenpeng Yin, and Dan Roth. 2020.
\newblock Paired representation learning for event and entity coreference.
\newblock \emph{arXiv preprint arXiv:2010.12808}.

\bibitem[{Zhang et~al.(2015)Zhang, Li, Ji, and Chang}]{zhang2015crossdocument}
Tongtao Zhang, Hongzhi Li, Heng Ji, and Shih-Fu Chang. 2015.
\newblock Cross-document event coreference resolution based on cross-media
  features.
\newblock In \emph{Proc. Conference on Empirical Methods in Natural Language
  Processing (EMNLP2015)}.

\end{thebibliography}
\bibliographystyle{acl_natbib}

% appendix has been moved to \newpage
\appendix
\section{Appendix}
% \subsection{Related Work}
% \input{appendix/appendix_related}
Section \ref{sec:feature_predictors} describes our symbolic feature predictors. Section \ref{sec:appendix_dataset} provides the details of the datasets we used. Section \ref{sec:appendix_hyperparameters} describes the hyperparameters and their value ranges that were explored. Section \ref{sec:reproducibility_checklist} presents our reproducibility checklist.

\subsection{Symbolic Feature Predictors} \label{sec:feature_predictors}
In an end-to-end setting, we train and use OneIE \cite{linetal2020joint} to identify event mentions along with their subtypes. For other symbolic features, we train a simple joint model. More specifically, given a document, our joint model first forms contextualized representations for the input tokens using SpanBERT \cite{Joshi2019SpanBERTIP}. Each event mention's representation is then defined as the average of the embeddings of the tokens in its trigger. After that, we feed the mentions' representations into classification heads for feature value prediction. Each classification head is a standard multi-layer feedforward network with softmax output units.

The event detection performance of OneIE on the test set of ACE 2005 is 74.7 (Type-F1 score). OneIE’s performance on KBP 2016 is 55.20 (Type-F1 score). For reference, the performance of the event detection component of $\text{E}^3\text{C}$ \cite{Lu2020EndtoEndNE} on KBP 2016 is 55.38 (Type-F1 score).

\subsection{Datasets Description}\label{sec:appendix_dataset}
In this work, we use two English within-document coreference datasets: ACE 2005 and KBP 2016. The ACE 2005 English corpus contains fine-grained event annotations for 599 articles from a variety of sources. We use the same split as that stated in \cite{chenetal2015event}, where there are 529/30/40 documents in the train/dev/test split. In ACE, a strict notion of event coreference is adopted, which requires two event mentions to be coreferential if and only if they had the same agent(s), patient(s), time, and location. For KBP 2016, we follow the setup of \cite{Lu2017LearningAS}, where there are 648 documents that can be used for training and 169 documents for testing. We train our model on 509 documents randomly chosen from the training documents and tune parameters on the remaining 139 training documents. Different from ACE, KBP adopts a more relaxed definition of event coreference, where two event mentions can be coreferent as long as they intuitively refer to the same real-world event. Table \ref{tab:dataset-statistics} summarizes the basic statistics of the datasets.

\renewcommand{\arraystretch}{1.15}
\begin{table}[!t]
\centering
\resizebox{\linewidth}{!}{%
\begin{tabular}{c|c|c|c}
\hline
Dataset & Train (\# Docs) & Dev (\# Docs) & Test (\# Docs) \\\hline
ACE-2005 & 529 & 30 & 40 \\
KBP 2016 & 509 & 139 & 169\\\hline
\end{tabular}
}
\caption{Basic statistics of the datasets.}
\label{tab:dataset-statistics}
\end{table}

\subsection{Hyperparameters} \label{sec:appendix_hyperparameters}
We use SpanBERT (spanbert-base-cased) as the Transformer encoder \cite{Wolf2019HuggingFacesTS,Joshi2019SpanBERTIP}. We did hyperparameter tuning using the datasets' dev sets. For all the experiments, we pick the model which achieves the best AVG score on the dev set, and then evaluate it on the test set. For each of our models, two different learning rates are used, one for the lower pretrained Transformer encoder and one for the upper layer. The optimal hyperparameter values are variant-specific, and we experimented with the following range of possible values: $\{8, 16\}$ for batch size, $\{$3e-5, 4e-5, 5e-5$\}$ for lower learning rate, $\{$1e-4, 2.5e-4, 5e-4$\}$ for upper learning rate, and $\{50, 100\}$ for number of training epochs. Table \ref{tab:noise_params} shows the value of $\epsilon$ we used for each symbolic feature type. In general, the larger the discrepancy between the train and test accuracies, the larger the value of $\epsilon$.

\renewcommand{\arraystretch}{1.25}
\begin{table}[!h]
\tiny
\centering
\resizebox{\linewidth}{!}{%
\begin{tabular}{|l|l|c|c|}
\hline
Dataset & Features & \begin{tabular}[c]{@{}c@{}}$\epsilon$ (predicted \\ mentions)\end{tabular} & \begin{tabular}[c]{@{}c@{}}$\epsilon$ (gold \\ mentions)\end{tabular} \\ \hline
\multirow{5}{*}{ACE 2005} & Type & 0.00 & 0.10 \\ \cline{2-4} 
 & Polarity & 0.00 & 0.02 \\ \cline{2-4} 
 & Modality & 0.15 & 0.20 \\  \cline{2-4} 
 & Genericity & 0.15 & 0.20 \\ \cline{2-4} 
 & Tense & 0.25 & 0.30 \\\hline
\multirow{2}{*}{KBP 2016} & Type & 0.05 & - \\ \cline{2-4} 
 & Realis & 0.10 & - \\ \hline
\end{tabular}%
}
\caption{The value of $\epsilon$ for each feature type.}
\label{tab:noise_params}
\end{table}

% This is because our preliminary experiments suggest that SpanBERT outperforms BERT

\subsection{Reproducibility Checklist} \label{sec:reproducibility_checklist}
We present the reproducibility information of the paper. Due to license reason, we cannot provide downloadable links for ACE 2005 and KBP 2016.

\paragraph{Implementation Dependencies Libraries} Pytorch 1.6.0 \cite{Paszke2019PyTorchAI}, Transformers 3.0.2 \cite{wolfetal2020transformers}, Numpy 1.19.1 \cite{Harris2020ArrayPW}, CUDA 10.2.

\paragraph{Computing Infrastructure} The experiments were conducted on a server with Intel(R) Xeon(R) Gold 5120 CPU @ 2.20GHz and NVIDIA Tesla V100 GPUs. The allocated RAM is 187G. GPU memory is 16G.

\paragraph{Average Runtime} Table \ref{tab:average_runtime} shows the estimated average run time of our full model.

\begin{table}[!ht]
\centering
\resizebox{\linewidth}{!}{%
\begin{tabular}{c|c|c|c}
Dataset & \begin{tabular}[c]{@{}c@{}}One Training \\ Epoch\end{tabular} & \begin{tabular}[c]{@{}c@{}}Evaluation \\ (Dev Set)\end{tabular} & \begin{tabular}[c]{@{}c@{}}Evaluation \\ (Test Set)\end{tabular} \\\hline
ACE 2005 & 65.5 seconds & 2.3 seconds & 2.4 seconds \\\hline
KBP 2016 & 103.6 seconds & 10.7 seconds & 11.8 seconds
\end{tabular}%
}
\caption{Estimated average runtime of our full model.}
\label{tab:average_runtime}
\end{table}

\paragraph{Number of Model Parameters} The number of parameters in a baseline model is about 109.7M parameters. The number of parameters in a full model trained on the KBP 2016 dataset is about 111.4M parameters. The number of parameters in a full model trained on the ACE 2005 dataset is about 113.8M parameters.

\paragraph{Hyperparameters of Best-Performing Models} Table \ref{tab:best_params} summarizes the hyperparameter configurations of best-performing models. Note that Table \ref{tab:noise_params} already showed the hyperparameters used for the noisy training method.

\begin{table}[!h]
\centering
\resizebox{\linewidth}{!}{%
\begin{tabular}{c|c|c|c}
Hyperparameters & \begin{tabular}[c]{@{}c@{}}ACE 2005 \\ (end-to-end)\end{tabular} & \begin{tabular}[c]{@{}c@{}}KBP 2016 \\ (end-to-end)\end{tabular} & \begin{tabular}[c]{@{}c@{}}ACE 2005 \\ (gold mentions)\end{tabular} \\\hline
Symbolic Features Used & All (CDGM) & All (CDGM) & All (CDGM) \\\hline
Noisy Training & Yes & Yes & Yes \\\hline
Lower Learning Rate & 4e-5 & 5e-5 & 5e-5\\\hline
Upper Learning Rate & 2.5e-4 & 5e-4 & 5e-4 \\\hline
Batch Size & 16 & 8 & 8 \\\hline
Number Epochs & 100 & 50 & 50\\\hline
\end{tabular}%
}
\caption{Hyperparameters for best-performing models (refer to Table \ref{tab:noise_params} for the hyperparameters used for the noisy training method).}
\label{tab:best_params}
\end{table}

\paragraph{Expected Validation Performance} We repeat training five times for each best-performing model. We show the average validation performance in Table \ref{tab:average_validation}. Our validation scores on KBP 2016 are not comparable to that of \cite{Lu2020EndtoEndNE}, because we split the original 648 training documents into the final train set and dev set randomly. We still use the same test set. For each best-performing model, we report the test performance of the checkpoint with the best AVG score in the main paper.

\begin{table}[!ht]
\centering
\resizebox{\linewidth}{!}{%
\begin{tabular}{c|c|c|}
Dataset & Avg. CoNLL score & Avg. AVG score \\\hline
\begin{tabular}[c]{@{}c@{}}ACE 2005 \\ (end-to-end)\end{tabular} & 60.20 & 58.80 \\\hline
\begin{tabular}[c]{@{}c@{}}KBP 2016 \\ (end-to-end)\end{tabular} & 54.86 & 48.83 \\\hline
\begin{tabular}[c]{@{}c@{}}ACE 2005 \\ (gold mentions)\end{tabular} & 81.9 & 83.02 \\\hline
\end{tabular}%
}
\caption{Average validation performance.}
\label{tab:average_validation}
\end{table}

\end{document}